\documentclass[letterpaper, 10pt, conference]{ieeeconf}

\IEEEoverridecommandlockouts
\usepackage[utf8]{inputenc}

\IEEEoverridecommandlockouts                              
\usepackage{times}

\usepackage{multirow}
\usepackage{xspace}
\usepackage{comment}
\usepackage{graphicx}
\usepackage[inkscapelatex=false]{svg}

\usepackage{xcolor}
\definecolor{gg}{RGB}{215, 25, 28}

\usepackage{todonotes}
\usepackage{amsmath}
\usepackage{amssymb}
\usepackage{amsthm}
\usepackage[flushleft]{threeparttable}
\usepackage{textcomp}

\newtheoremstyle{main}
                {1em}                                                
                {1em}                                                
                {\normalfont}                                        
                {0pt}                                                
                {\scshape}                                           
                {\\*}                                                
                {2pt}                                                
                {\thmname{#1}\thmnumber{ #2}: \thmnote{\itshape #3}} 

\theoremstyle{main}

\usepackage[linesnumbered,ruled,noend]{algorithm2e}

\usepackage[
  activate   = {true},
  protrusion = true,
  expansion  = true,
  kerning    = true,
  spacing    = true,
  tracking   = false,
  auto       = true,
  selected   = true,
  factor     = 1000,
  stretch    = 15,
  shrink     = 15,
]{microtype}

\makeatletter
\let\NAT@parse\undefined
\makeatother
\usepackage[pdfa,colorlinks,bookmarksopen,bookmarksnumbered,allcolors=gg]{hyperref}
\usepackage{cite}

\newcommand{\flame}{\textsc{flame}\xspace}
\newcommand{\fire}{\textsc{fire}\xspace}

\newcommand{\C}{\ensuremath{\mathcal{C}}\xspace}
\newcommand{\D}{\ensuremath{\mathcal{DS}}\xspace}
\newcommand{\DB}{\ensuremath{\mathcal{DB}}\xspace}
\newcommand{\SIM}{\ensuremath{\mathcal{S}}\xspace}
\newcommand{\DSIM}{\ensuremath{\mathcal{NS}}\xspace}
\newcommand{\W}{\ensuremath{\mathcal{W}}\xspace}
\newcommand{\M}{\ensuremath{\mathcal{M}}\xspace}
\newcommand{\cspace}{\mbox{\ensuremath{\mathcal{C}}-space}\xspace}
\newcommand{\samples}{\mbox{``critical samples''}\xspace}
\newcommand{\samplesp}{\mbox{``critical samples.''}\xspace}

\newcommand{\regions}{\mbox{``challenging regions''}\xspace}

\newcommand{\norm}[1]{\left\lVert#1\right\rVert}

\newcommand{\rrtc}{\textsc{rrtc}\xspace}
\newcommand{\rrtcd}{\textsc{rrtc-default}\xspace}
\newcommand{\rrtct}{\textsc{rrtc-tuned}\xspace}
\newcommand{\biest}{\textsc{biest}\xspace}
\newcommand{\biestd}{\textsc{biest-default}\xspace}
\newcommand{\biestt}{\textsc{biest-tuned}\xspace}
\newcommand{\static}{\textsc{static}\xspace}
\newcommand{\mpnet}{\textsc{mpnet-smp}\xspace}
\newcommand{\uni}{\textsc{uniform}\xspace}

\newcommand{\smalld}{\mbox{``Small-Shelf''}\xspace}
\newcommand{\talld}{\mbox{``Tall-Shelf''}\xspace}
\newcommand{\thind}{\mbox{``Thin-Shelf''}\xspace}
\newcommand{\caged}{\mbox{``Cage''}\xspace}
\newcommand{\tabled}{\mbox{``Table''}\xspace}

\newcommand{\threed}{\textsc{3d}\xspace}

\newcommand{\dof}{\textsc{dof}\xspace}

\newcommand{\lp}{\ensuremath{\ell}\xspace}
\newcommand{\LP}{\ensuremath{\mathcal{L}}\xspace}

\newcommand{\lw}{\ensuremath{lw}\xspace}
\newcommand{\xs}{\ensuremath{x_\textsc{start}}\xspace}
\newcommand{\xg}{\ensuremath{x_\textsc{goal}}\xspace}
\newcommand{\xp}{\ensuremath{x_\textsc{proj}}\xspace}
\newcommand{\xt}{\ensuremath{x_\textsc{target}}\xspace}
\newcommand{\sfun}{\textsc{sim}\xspace}
\newcommand{\prj}{{\pi}\xspace}
\newcommand{\Prj}{{\Pi}\xspace}
\newcommand{\xps}{\ensuremath{v}\xspace}
\newcommand{\xpst}{\ensuremath{\tilde{\xps}}\xspace}

\newcommand{\obs}{\text{obs}\xspace}
\newcommand{\free}{\text{free}\xspace}

\newcommand{\pv}{\ensuremath{p}\xspace}
\newcommand{\ui}{\ensuremath{[0, 1]}\xspace}

\usepackage{todonotes}

\title{\LARGE \bf
    Learning to Retrieve Relevant Experiences for Motion Planning
}

\author{Constantinos Chamzas, Aedan Cullen, Anshumali Shrivastava,  Lydia E. Kavraki 
\thanks{All authors are affiliated with the Department of Computer Science, Rice University, Houston TX, USA {\tt\small\{chamzas, aedan, anshumali, kavraki\}@rice.edu}. This work was supported in part by NSF 1718478, NSF-GRFP 1842494 and Rice University Funds.}
}

\begin{document}

\maketitle
\thispagestyle{empty}
\pagestyle{empty}

\begin{abstract}
Recent work has demonstrated that motion planners' performance can be significantly improved by retrieving past experiences from a database.
Typically, the experience database is queried for past similar problems using a similarity function defined over the motion planning problems.
However, to date, most works rely on simple hand-crafted similarity functions and fail to generalize outside their corresponding training dataset.
To address this limitation, we propose (\fire), a framework that extracts local representations of planning problems and learns a similarity function over them.
To generate the training data we introduce a novel self-supervised method that identifies similar and dissimilar pairs of local primitives from past solution paths.
With these pairs, a Siamese network is trained with the contrastive loss and the similarity function is realized in the network's latent space.
We evaluate \fire on an 8-\dof manipulator in five categories of motion planning problems with sensed environments.
Our experiments show that \fire retrieves relevant experiences which can informatively guide sampling-based planners even in problems outside its training distribution, outperforming other baselines.
\end{abstract}

\section{Introduction}
Motion planning is used in real-time autonomous vehicles~\cite{Kuwata2009}, manipulators in dynamic environments~\cite{Murray2016},
and as a subroutine in planners for complex missions (e.g. task and motion planning~\cite{Dantam2018}), all of which rely heavily on efficiency.
However, motion planning is still challenging, especially for high-dimensional systems~\cite{Canny1988}.
Sampling-based planners \cite{kavraki1996, Hsu1999, LaValle2000} are a class of motion planning algorithms that have found widespread adoption in the planning community.
Although significant progress has been made over the years, planning is still computationally expensive~\cite{Salzman2019}, hindering the adoption of robotic solutions.
Thus, to endow robots with real-time capabilities, faster motion planning algorithms are necessary.

\begin{figure}[ht!]
  \centering

  \vspace{-1em}
  \includegraphics[width=0.9\linewidth]{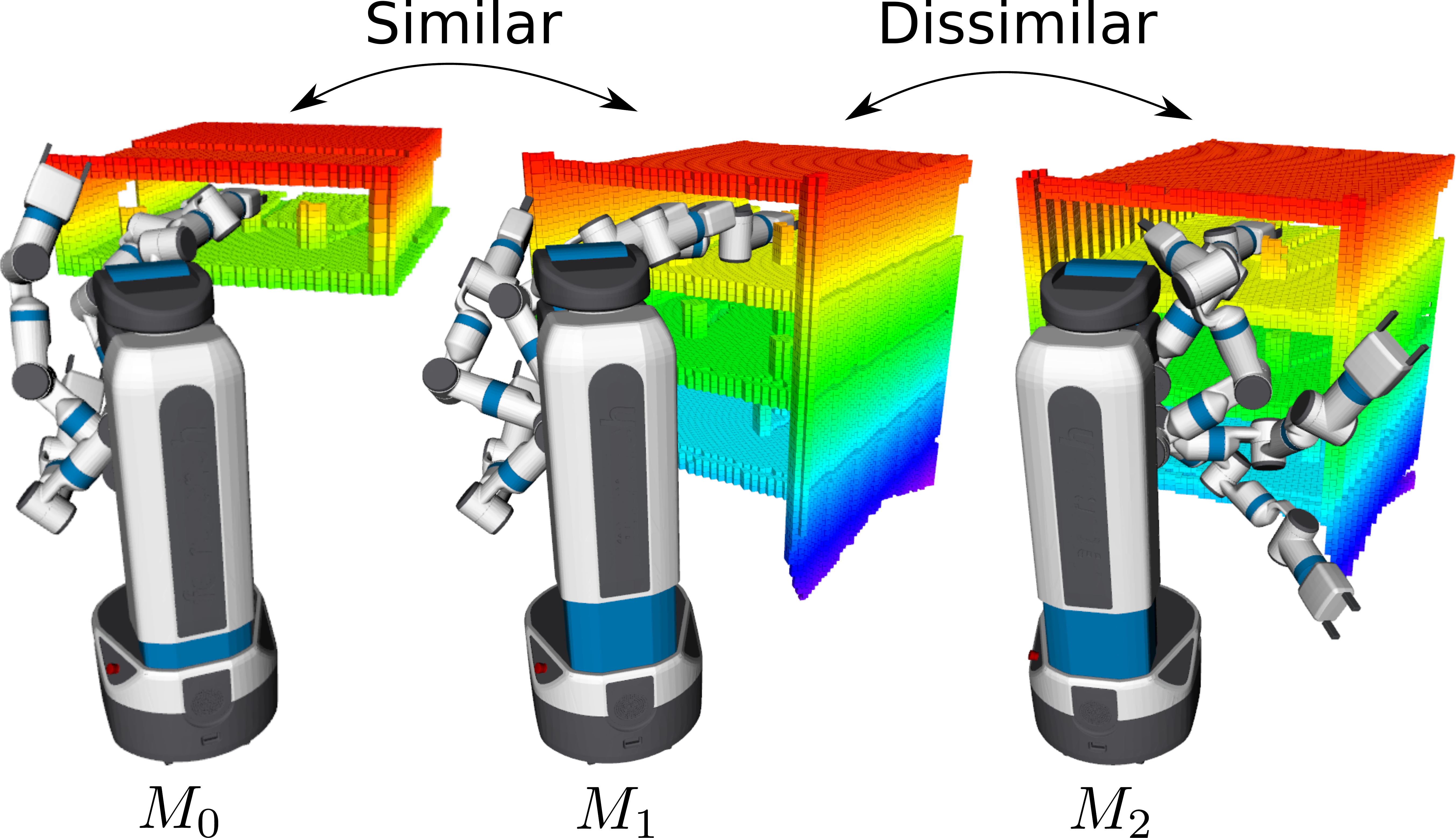}  
  \caption{Three example problems $M_0, M_1, M_2$ where the robot is tasked with picking one object from a shelf starting from the same tuck (home) configuration (not shown for visual clarity).
      The motion planning problems $M_0$ and $M_1$ have similar solution paths even though their workspaces are visually different.
      On the other hand, visually similar workspaces such as $M_1$ and $M_2$ can have different solution paths for subtle reasons (e.g. slightly different goals, obstacle arrangements, and robot base orientation).
       } 
  
\label{fig:intro}
\end{figure}

A promising avenue is to guide planning by leveraging the past experiences of a robot. 
Several methods have shown that storing and retrieving experiences~\cite{Jetchev2013, Coleman2015thunder} can significantly improve motion planners' efficiency. These methods have focused on what to store and how to adapt/repair it for the current situation, but not on how to retrieve the most relevant experiences, defaulting to simple similarity functions. 
In other words, little emphasis has been placed on finding suitable functions that quantify the similarity of motion planning problems, limiting the generalizability of retrieval-based methods outside their training dataset. 

In this context, for similar motion planning problems or subproblems, the solution path of one can be used to expedite the search when solving the other. 
Capturing this notion of similarity is the core investigation of this work.
Designing a good similarity function is very challenging for motion planning problems.
For example, in \autoref{fig:intro} two visually dissimilar workspaces $M_0$, $M_1$ have similar solution paths while visually similar workspaces $M_1$ and $M_2$ have different solution paths. A good similarity function should capture the commonalities between $M_0$ and $M_1$ while still distinguishing between $M_1$ and $M_2$. These problems are part of the \talld dataset described in \autoref{sec:exp}.

To address this problem we propose \textbf{F}ast retr\textbf{I}eval of \textbf{R}elevant \textbf{E}xperiences (\fire).
As detailed in \autoref{sec:meth}, \fire extracts suitable local representations, called local primitives, from previous problems.
\fire finds pairs of similar and dissimilar local primitives using a self-supervised method.
With these pairs, a similarity function is learned which can be used to retrieve relevant experiences and guide a motion planner. 
We demonstrate the effectiveness of \fire with an 8-\dof mobile manipulator in five categories of diverse problems with sensed environments as shown (\autoref{fig:intro}).
Through our experiments (\autoref{sec:exp}) we show that \fire generalizes better outside its training dataset even with less data, and is faster in terms of planning time than prior work.
The implementation of \fire and the generated datasets are open-source \footnote{\mbox{\url{https://github.com/KavrakiLab/pyre}}}.

Overall, the main contributions of this work lie in 1) defining suitable local representations of motion planning problems, 2) learning a similarity function over them, and 3) applying it in the motion planning problem through our new framework. 
Although \fire is tailored to retrieval frameworks that use local features and biased sampling distributions~\cite{Chamzas2019, Chamzas2021} we believe it could be easily adapted to work with other retrieval-based methods~\cite{Lien2009, Tang2019, Merkt2020}.

\section{Problem Description and Notation}\label{sec:form} 

\textit{Feasible Path Planning}:
Consider a robot in a workspace \W.
A configuration of the robot $x$ is a point in the configuration space (\cspace), $x \in \C$.
Obstacles in the workspace induce \cspace obstacles $X_\obs \subset \C$.
The set of configurations that are not in collision is denoted by $X_\free = \C - X_\obs$.
We are interested in finding a path \pv, from $\xs \in X_\free$ to $\xg \in X_\free$, as a continuous map  with $\pv(0) = \xs,~\pv(1) = \xg$ such that for all $t \in \ui$, $\pv(t) \in X_\free$.
We denote the motion planning problem by \mbox{$ \M =(\xs, \xg, \W)$}.

\textit{``Challenging Regions'' and ``Critical Samples''}:
In this work, we are concerned with planning for high-dimensional robotic manipulators, and focus on sampling-based motion planners. 
A common theme in learning-based approaches is to produce configurations in \cspace regions with low visibility~\cite{Hsu2003}, which are the main bottleneck for sampling-based motion planners~\cite{Hsu2006}.
We denote these \regions, and configurations inside them \samplesp

\textit{Retrieval-Based Learning for Motion Planning}:
Given a dataset \mbox{$\D=\{\M^i:\pv^i\}^N_{i=1}$} of past problems \M and their feasible paths \pv, retrieval-based methods extract information from \D and store it in a database denoted \DB.
In this context, \DB is a structure that contains $\langle key:value \rangle$ entries, with the experiences (values) being \samplesp
The indices (keys) of the database are local primitives denoted by $\lp \in \LP$, where \LP is the space of local primitives.
Each local primitive includes local workspace information~\cite{Chamzas2021} along with \xs , \xg information (as defined in~\autoref{sec:lp}).
This work aims to learn a suitable similarity function  \mbox{$\sfun:\LP\times\LP \rightarrow \{0,1\} $} over the local primitives in order to retrieve relevant \samples for a given problem \M.

\section{Related work}\label{sec:related}
Over the years many techniques have been proposed to guide sampling-based motion planners. 
Many examples use heuristics to bias sampling, such as Bridge sampling~\cite{Hsu2006}, Gaussian sampling~\cite{Boor1999}, Medial-Axis sampling~\cite{Lien2003},
and workspace-based sampling~\cite{Kurniawati2008}.
However, these predefined heuristics may or may not apply in different situations.

Thus, a growing number of works attempt to learn how to guide planning by utilizing past solutions to motion planning problems.
One set of methods learns interesting regions in \W~\cite{Zucker2008, Molina2020Link} but requires an inverse kinematics solver to infer samples in \regions.
A similar class of methods directly computes relevant configurations in \C from a motion planning problem \M using a neural network.
For example, some methods train a conditional variational autoencoder to reconstruct samples from previous paths~\cite{Ichter2018}
or \regions~\cite{Ichter2020LocalCrit, Kumar2019, Jenamani2020LocalCrit}.
The authors of~\cite{Patil2019PredictionOB, Terasawa20203d} use a \threed CNN to sample in \regions, while~\cite{Qureshi2020motion, Tamar2019, Chen2020} use neural networks as motion planners. 

Although these methods have shown some promising results, mapping \M to paths or \regions in \C is hard in high-dimensional problems.
Motion planning is sensitive to input; small changes in $\W$, $\xs$, or $\xg$ can drastically alter the resulting solution~\cite{Tang2019, Chamzas2021, Farber2003}.
Furthermore, this mapping is usually multi-modal, since a motion planning problem may have multiple solution paths or multiple disjoint \regions~\cite{Merkt2020, Rice2020Multihomotopy}.

For these reasons, some approaches have adopted retrieval-based methods, also known as library-~\cite{Stolle2007Transfer} or memory-based~\cite{Lembono2020} methods.
Such methods typically store in memory a database $\DB$ and retrieve relevant information in the form of paths~\cite{Berenson2012lightning, Pairet2021} or sampling distributions~\cite{Chamzas2019,Finney2007Partial} based on a similarity function over \M.
These methods naturally apply to multi-modal problems, since for similar or identical \M multiple outputs can be retrieved.
Another advantage of these methods is that they are incremental since new experiences can simply be added to the database \DB.
The main challenge lies in constructing a good similarity function over \M.

Defining a similarity function is challenging because \M contains heterogeneous parameters; $\xs, \xg \in \C$ while $\W$ is a 3D representation. 
Some approaches do not use a similarity function but learn problem invariants \cite{Iversen2016Kernel, Lehner2017}, others construct the similarity only over $\xs$ and $\xg$~\cite{Coleman2015thunder, Berenson2012lightning}, and some construct it only over \W~\cite{Lien2009, Chamzas2021}.
In~\cite{Chamzas2021} a hand-crafted similarity function over local workspaces is defined, while~\cite{Lien2009} defines workspace similarity based on geometric deformation of obstacles.
Most similarly to our work, \cite{Jetchev2013} learned a similarity function over $\xs$, $\xg$, and \W using a weighted combination of global workspace features.
In contrast, our work uses local features and leverages latent space representations obtained from neural networks.

Learning similarity functions~\cite{hoffer2015deep} in the latent space has been successfully employed in computer-vision tasks, such as
image classification~\cite{vinyals2016matching} and 3D object classification~\cite{zeng20163dmatch}.
Our work is inspired by these methods, and applies similar metric learning methods to the motion planning problem.

\section{Methodology}\label{sec:meth}
\begin{figure}
  \centering
  \vspace{-0.5em}
  \includegraphics[width=0.8\linewidth]{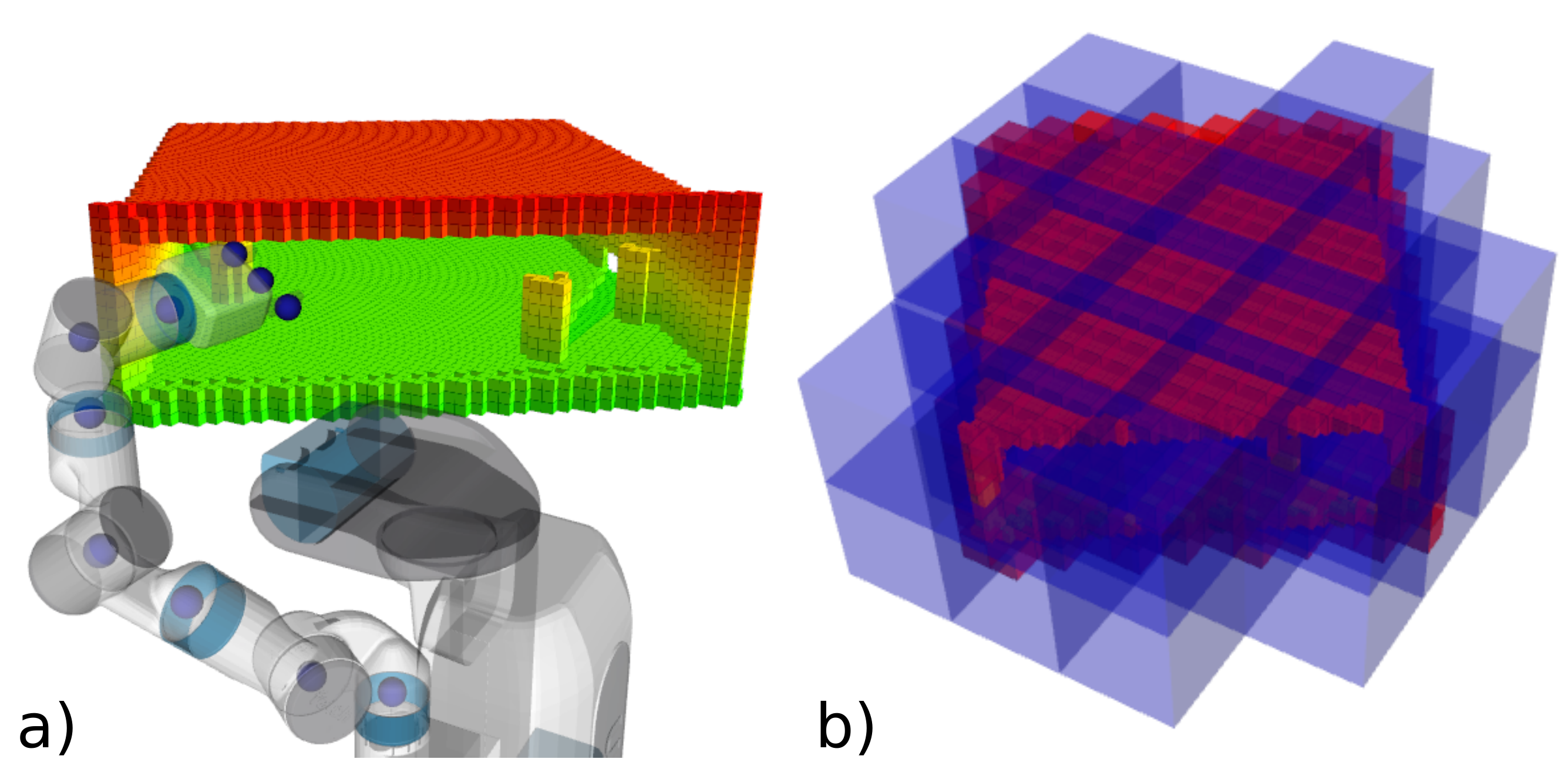}  
  \vspace{-0.5em}
  \caption{ \textbf{a)} The blue dots depict the 10 projections defined on the arm and gripper of the Fetch robot.
      Each blue dot is one projection point $\pi(x)_p \in \mathbb{R}^{3}$ of $x \in \C$.
      Specifically, each robotic link of the arm+gripper was used as a projection, as described in its \texttt{urdf}.
  \textbf{b)} Examples of local occupancy grids and their position in space  derived from the sensed scene ($\lw = (b, \xps$)).
Note that only non-empty local occupancy grids are generated. }
  \label{fig:proj_octo}
\end{figure}

We propose \fire, a framework that learns a similarity function to retrieve relevant experiences from a database in the form of \samples.
In \autoref{sec:lp} we formulate the local primitives which are the input to the similarity function, and we extract them from past problems in \autoref{sec:cr}.
Then, we describe how to generate similar and dissimilar local primitives~\mbox{(\autoref{sec:gen})}.
In \autoref{sec:learn}, we train a Siamese network by minimizing the contrastive loss of the local primitive pairs and realize the similarity function in the learned latent space.
Finally, \autoref{sec:ret} explains how the similarity function can guide a sampling-based planner.   

\subsection{Local primitives}\label{sec:lp}
First, we define a set of projections $ \prj (x): \C \rightarrow \mathbb{R}^3$ used to extract and compare local primitives. 
Each configuration $x$ is projected to multiple points in \W and stacked as a vector
$$\Prj(x)= [\prj_1(x), \prj_{2}(x), \ldots,  \prj_{P}(x)] \in \mathbb{R}^{3\times P}$$
where $P$ is the number of projections.
\autoref{fig:proj_octo}a) shows the 10 projections on the Fetch  which we used. Specifically, we used the link frames of the arm+gripper from the Fetch \cite{Wise2016} \texttt{urdf}.
Projections have often been used to guide motion planners~\cite{Orthey2018quotient} and specifying them is often a research problem in itself, albeit outside the scope of this work.
  
Now we define the local primitives \lp, which include a local \threed occupancy grid and its position \lw \cite{Chamzas2021} along with some auxiliary \cspace information \xt and \xp:

\begin{align*}
    \lp =  [\lw, \xt, \xp ] 
\end{align*}

More specifically, $\lw = (b, \xps)$ where  
$b \in \{0,1\}^{64}$ is a \mbox{64-bit} binary vector that represents a (4x4x4) local occupancy grid and $\xps \in \mathbb{R}^3$ is the center position of the grid. 
Examples of \lw are shown in \autoref{fig:proj_octo}b.
The variable $\xt \in \C$ is either \xs or \xg, depending on the situation as explained in \autoref{alg:1} and \autoref{sec:ret}.
Finally, we calculate \xp from  \xt and the center position \xps of \lw.
We project \xt to $P$ points in the workspace $\Prj(\xt)  \in \mathbb{R}^{3\times P}$ and then aggregate all the distances between the $P$ points and \xps to calculate \xp:
\begin{align*}
     \xp= [\norm{\xps - \prj_1(\xt)}, \ldots , \norm{\xps - \prj_P(\xt)}]  \in \mathbb{R}^{P} 
\end{align*}
The variable \xp serves as an interleaved representation of \xt and \lw and was empirically validated to improve the latent space structure. 

\subsection{Creating the experience database} \label{sec:cr}

\autoref{alg:1} describes how to create the experience database \DB from \mbox{$\D=\{(\xs, \xg, W)^i:\pv^i\}^N_{i=1}$} by associating each local primitive with a configuration from a solution path.

First, the paths are shortcutted~\cite{Raveh2011} to remove redundant nodes not in \regions (\autoref{alg:1:short} in \autoref{alg:1}) and keep only \samples.
Finding \samples is still an open research problem~\cite{Ichter2020LocalCrit, Molina2020Link, Chamzas2021} but this simple shortcutting heuristic has been used previously in~\cite{Chamzas2019, Iversen2016Kernel}.

Next, \verb|TARGET| (\autoref{alg:1:target} in \autoref{alg:1}) samples near \xs and \xg and chooses the one which yielded the most in-collision samples with the workspace.
This aims to create the same local representation for motion plans with the same solution path but swapped \xs and \xg.
Consider for example the task in \autoref{fig:intro}, where the robot plans from the home (\xs) to a grasp configuration (\xg).
The same solution path applies for planning between the grasp configuration (\xs) back to the tuck configuration (\xg). Thus, to ensure that both plans have the same local representations \verb|TARGET| should choose the same configuration as \xt (e.g. the grasp configuration).
We then decompose the workspace to local occupancy grids (\autoref{alg:1:decompose} in \autoref{alg:1}) by traversing the octomap tree similarly to~\cite{Chamzas2021}.

\begin{algorithm}
  \caption{Creating the experience database \label{alg:1}}

  \SetKwInOut{Input}{Input}
  \SetKwInOut{Output}{Output}
  \SetKwFunction{TARGET}{TARGET}
  \SetKwFunction{SHORT}{SHORTCUT}
  \SetKwFunction{CONTAINS}{CONTAINS}
  \SetKwFunction{NEXT}{NEXT}
  \SetKwFunction{PREVIOUS}{PREV}

  \Input{MP problem $\W ,\xs ,\xg$ Path \pv}
  \Output{Database \DB}
   Shortcut $\pv^{\prime} = \SHORT(\pv)$ \\ \label{alg:1:short}
   Find target \xt $\gets$ \TARGET(\xg, \xs)\\ \label{alg:1:target}
   Decompose $\W$ to $\mathcal{LW} \gets \{\lw_{1}, \ldots, \lw_{M}\}$ \label{alg:1:decompose} \\
   \ForEach{$\lw \in \mathcal{LW}$}
   {
       \ForEach{$x \in \pv^{\prime}$}
       { 
           \ForEach{$\prj \in \Prj $}
           { 
               \label{alg:1:contains}
               \If{\CONTAINS (\lw, $\prj(x)$)} 
               {
                   $\xp \gets |\xpst - \Prj(\xt)|$ \\
                   $\lp \gets [\lw, \xt,\xp]$ \\
                   $x^{n} \gets \NEXT(x,\pv)$ \\  
                   $x^{p} \gets \PREVIOUS(x,\pv)$ \\  
                   Insert $\langle \lp: x^{p}, x,  x^{n} \rangle$ in \DB \label{alg:1:store}
               }
           }
        }
   }
\Return \DB
\end{algorithm}

Afterward, we iterate over the configurations in each path, the local occupancy grids, and the projections.      
The subroutine \verb|CONTAINS| associates each configuration with its relevant regions in the workspace.
\verb|CONTAINS| checks for every projection $\prj(x)_p \in \mathbb{R}^{3}$ of the configuration $x$ if it is contained in the bounding box of an occupancy grid; if so we store the local primitive \lp along with the critical $x$, the previous waypoint configuration $x^p$, and the next waypoint configuration $x^n$ in \DB.
The previous and next configurations are only used to help us create similar pairs as described in \autoref{alg:2} and are not part of the retrieved experience.
\subsection{Creating a dataset of similar pairs}\label{sec:gen}
\autoref{alg:2} describes a novel method to create a dataset of similar pairs of local primitives over which to learn the similarity function. This is the key problem investigated in this paper.

Given a database \DB, we iterate over all pairs of local primitives and perform the following checks.
First, the subroutine \verb|SAME_PROJ| checks if the two local primitives were generated by the same projection (\autoref{alg:2:proj} in \autoref{alg:2}).
Then we check whether the centers $v$ of the local occupancy grids are close enough in \W  (\autoref{alg:2:check1} in \autoref{alg:2}) and whether the stored configurations are also close enough in \cspace (\autoref{alg:2:check2} in \autoref{alg:2}).
The variable $lw_{side}$ is the length of the side of the local occupancy bounding box \lw.

Finally~(\autoref{alg:2:check2} in \autoref{alg:2}) we sample up to N times $x^{near}_{j} \sim  \mathcal{N}(x_j, \sigma^2)$ until a configuration $x^{near}_{j}$ is found which passes the \verb|VALID| check.
The \verb|VALID| subroutine checks if $x^{near}_{j}$ can connect through a collision-free edge (in the full workspace \W of $\lp_i$) with the next $x^n_{i}$ and previous $x^p_{i}$ configuration of the local primitive $\lp_i$.
If such a configuration is found then we consider $\langle \lp_i,\lp_j \rangle$ similar and add them to \SIM. 
This procedure aims to discover local primitives whose \samples are good substitutes for one another by emulating how \samples are used to bias sampling during planning (\autoref{sec:ret}).
To generate dissimilar pairs we randomly choose local primitives from \DB and generate an equal number of dissimilar pairs.  We denote the set that includes these dissimilar pairs \DSIM.

\begin{algorithm}
  \caption{Creating a dataset of similar pairs \label{alg:2}}

  \SetKwInOut{Input}{Input}
  \SetKwInOut{Output}{Output}
  \SetKw{Break}{break}
  \SetKwFunction{CHECK}{VALID}
  \SetKwFunction{COMMON}{SAME\_PROJ}
  \SetKwFor{RepTimes}{repeat}{times}{end}

  \Input{Database \DB}
  \Output{Pairs of similar local primitives \SIM}
  \ForEach{$\langle \lp_i: x^p_{i}, x^{i},  x^{n}_{i} \rangle \in \DB$}
  {
      \ForEach{$\langle \lp_j: x^p_{j}, x_{j},  x^n_{j} \rangle \in \DB$}
      {
          \If{\COMMON{$\lp_i, \lp_j$}}
          {
              \label{alg:2:proj}
              \If{$\norm{\xps_j-\xps_j}_1 <= size_{\lw} $}
              {
                  \label{alg:2:check1}  
                  \If{$||x_i-x_j||< 10\sigma^2$}
                   {
                       \label{alg:2:check2}  
                       \RepTimes{N}
                       {
                           \label{alg:2:check3}  
                           $x^{near}_{j} \sim  \mathcal{N}(x_j, \sigma^2)$ \\
                           \If{\CHECK{$x^p_{i}, x^{near}_{j},  x^n_{i}$}}
                           {
                               \SIM $\gets \langle \lp_j, \lp_i \rangle$ \\
                               \Break \\
                           }
                       }

                   }
               }
           }
      }
  }
  \Return \SIM
\end{algorithm}

Note that~\autoref{alg:2} needs the \samples extracted from solution paths to find similar local primitives, and cannot be used as a similarity function when solving a new motion planning problem where only $W, \xg, \xs$ is available.

\subsection{Learning the similarity function}\label{sec:learn}
\begin{figure}
  \centering
  \vspace{1em}
  \includesvg[width=0.9\linewidth]{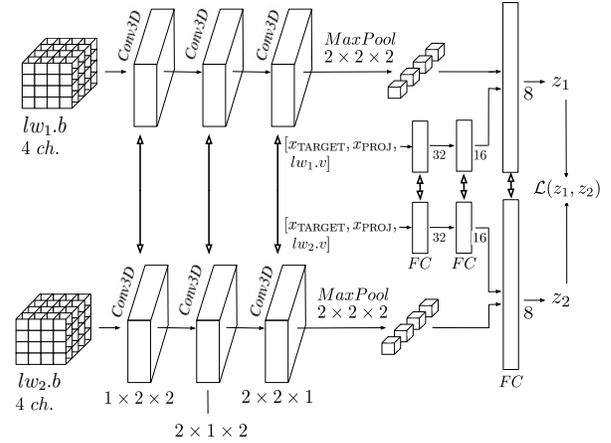}
  \vspace{-1em}
  \caption{The Siamese network architecture used. The activation function for all the layers was ReLU.
\textit{Conv3D} denotes a 3D convolutional layer,
\textit{MaxPool} takes the maximum value out of every subgrid, and \textit{FC} denotes a fully connected layer. The parameters of each layer are shown in the figure.
}
  \label{fig:siamese}
\end{figure}

The learned similarity function is realized in the latent space of a Siamese network.
A Siamese network~\cite{chicco2020siamese} is comprised of two identical encoder networks as shown in \autoref{fig:siamese}.
Each encoder maps \lp to a latent variable \mbox{$z \in \mathbb{R}^{8}$}.
The overall network is relatively small with around 3500 parameters, 
and was trained with the contrastive loss~\cite{hadsell2006dimensionality}: 
\begin{equation*}\label{eq:contrastive}
\mathcal{L}(\lp_i, \lp_j) =
\begin{cases}
    \max (0\xspace,d_m - \norm{z_i-z_j}^2) & \text{if }   \langle \lp_j, \lp_i \rangle \in \DSIM \\
            ||z_i - z_j||^2 & \text{if } \langle \lp_j, \lp_i \rangle \in \SIM
\end{cases}
\end{equation*}

This loss tries to bring local primitives that belong in \SIM~(similar) as close as possible in the latent space $Z$, while local primitives that belong in~\DSIM (dissimilar) must have at least a  margin distance $d_m=0.5$.
After having structured the latent space $Z$ the similarity function is defined as follows:

\begin{equation*}\label{eq:simfun}
\sfun(\lp_i, \lp_j) =
\begin{cases}
    1 & \text{if} \norm{z_i-z_j}^2< R\\
    0 & \text{ otherwise} 
\end{cases}
\end{equation*}

\noindent where $R=0.2d_m$ is the retrieval radius.
A lower retrieval radius than the margin distance $d_m$ must be used to avoid retrieving dissimilar pairs. 
After structuring the latent space $Z$ all the local primitives in \DB are projected to $Z$ and added in a \mbox{K-D tree}~\cite{bentley1975multidimensional} structure for fast retrieval.
Finding similar local primitives with \sfun is equivalent~\cite{balcan2008theory} to retrieving all the neighbors within radius $R$ in the latent space $Z$. 

\subsection{Retrieving relevant experiences} \label{sec:ret}

When solving a new problem \mbox{\M=(\xs,\xg, \W)} the new local primitives are created with the following procedure.
First, we extract the local occupancy grids from \W.
Then, for each local occupancy grid \lw we generate two local primitives: one with $\xt=\xs$ and one with $\xt=\xg$.
The value of $\xp$ is calculated from \xt and \lp as explained in \autoref{alg:1}.
Each created local primitive is projected to $Z$ and its neighbors within radius $R$ are retrieved, effectively obtaining their associated \samples from \DB.
Finally, similarly to~\cite{Chamzas2021}, we aggregate all the $K$ \samples and convert them to a Gaussian Mixture Model (\textsc{gmm}):
$$P(x|\M) = \frac{1}{K} \sum^{K}_{i=0} \mathcal{N}(x_i, \sigma^2)$$

The \textsc{gmm} can be used to bias the sampling of any sampling-based planner.
To keep the probabilistic completeness guarantees of sampling-based planners we sample from $P(x|\M)$ with probability $0< \lambda <1 $ and from a standard uniform distribution with probability $(1-\lambda)$. 
If the planner uses a local expansion strategy like \textsc{est}~\cite{Hsu1999} we simply sample from the mixtures that are within the local sampling radius.

\section{Experiments}\label{sec:exp}
\begin{figure}
  \centering
  \includesvg[width=\linewidth]{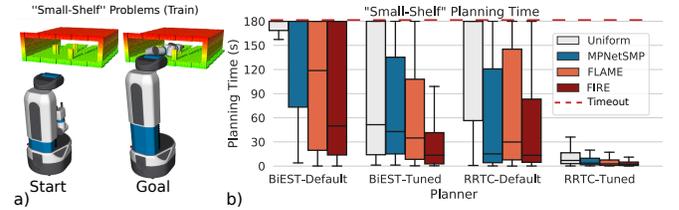}  
  \vspace{-1.5 em}
  \caption{\textbf{a)} An example problem from the \smalld dataset.
      We generate different problems by uniformly sampling the robot pose, the position of the obstacles, and the height of the shelf.
      This is similar to the \smalld used in~\cite{Chamzas2021} but the shelf is shorter, making it more challenging due to the narrow area the robot has to traverse.
      \textbf{b)} Planning time (including retrieval) with different underlying planners for 100 test examples from the \smalld dataset. The timeout was set to 180 seconds.}
  \label{fig:small}
\end{figure}

\begin{figure*}
  \vspace{1em}
  \centering
  \includegraphics[width=\linewidth]{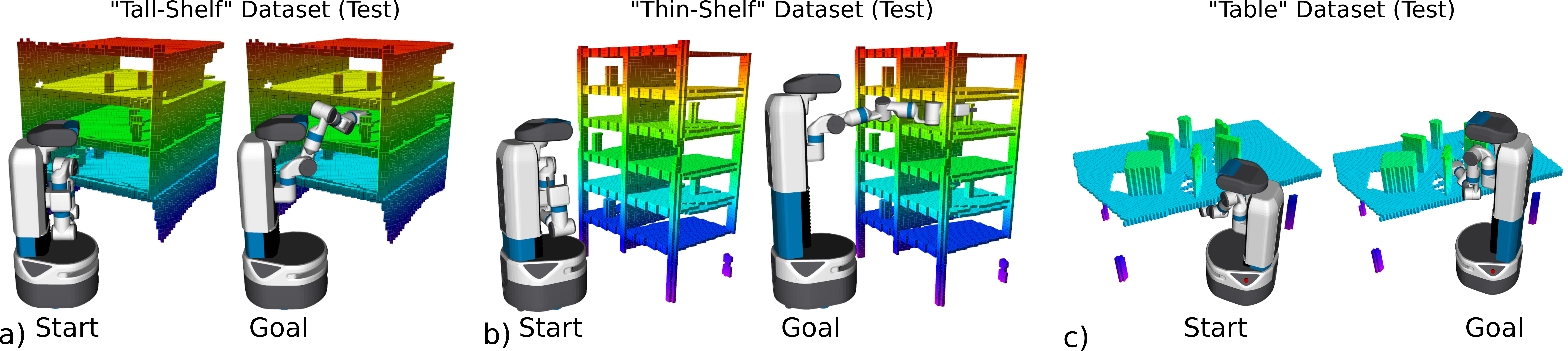}  
  \vspace{-1.75em}
  \caption{The three datasets used to test the evaluated methods. Different problems are generated similarly to \autoref{fig:small}.
      \textbf{a)} An example environment from the \talld dataset. The \talld is created by stacking the \smalld three times.
      \textbf{b)} An example environment from the \thind dataset. This is also a bookcase like \smalld and \talld, but the shelves are shorter and there is a divider, making it a much more challenging problem.
      \textbf{c)} An example environment from the \tabled dataset, which includes a table with several objects and is very different from the other datasets.}
  \label{fig:envs}
\end{figure*}

We demonstrate the effectiveness of the learned similarity function on five generated datasets with \textsc{MotionBenchMaker} \cite{Chamzas2022}. 
Each dataset contains an 8-\dof (arm+torso) Fetch robot~\cite{Wise2016} with a workspace represented by an octomap~\cite{Hornung2013}, performing a pick task as shown in \autoref{fig:small}a.
We consider this a realistic representation since point clouds can easily be obtained from a simple depth camera.
The five datasets generated were \smalld (\autoref{fig:small}a), \talld (\autoref{fig:envs}a), \thind (\autoref{fig:envs}b), \tabled (\autoref{fig:envs}c), and \caged (\autoref{fig:cage}a).
As shown in the figures, the starting configuration \xs for all datasets was a home (tuck) position, except for \tabled where \xs is a random configuration under the table.
The goal configuration \xg is an inverse kinematics (IK) solution placing the end-effector in a grasping pose relative to an object.
For the ``Shelf'' datasets, one object per shelf is grasped and it is always the one furthest back.
For \tabled and \caged the grasped object is shown in the figures.
We generate different motion planning problems similarly to~\cite{Chamzas2021} by uniformly sampling  poses for the robot base and scene objects.
Note that such variation generates highly diverse planning problems since even small changes in the positions of the obstacles relative to the robot drastically affect $X_\obs$ and the resulting \xg.

\begin{figure*}
  \centering
  \vspace{-0.5em}
  \includesvg[width=\linewidth]{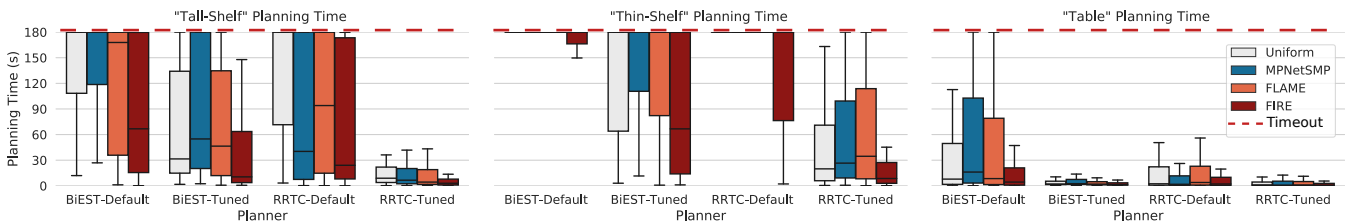}  
  \vspace{-2em}

  \caption{
  Planning time (including retrieval) when testing in the three datasets shown in \autoref{fig:envs}.
  All of the methods are only trained with the \smalld dataset. The timeout was set to 180 seconds.
  } 
  \vspace{-1em}
  \label{fig:test}
\end{figure*}

All evaluated methods produce biased samples in \C which can guide any sampling-based motion planner.
We evaluated these methods within \textsc{rrt}-connect (\rrtc)~\cite{Kuffner2000} and bidirectional \textsc{est} (\biest)~\cite{Hsu1999}, implemented in the Open Motion Planning Library (\textsc{ompl})~\cite{Sucan2012}.
Additionally, we considered two versions of each planner: one with default \textsc{ompl} parameters (\rrtcd and \biestd) and one with a tuned range parameter (\rrtct and \biestt) found by a parameter sweep over a diverse set of problems.
In our experiments we compare \fire with the following methods: 
\begin{itemize}
\item \uni : Default uniform sampling of the \cspace.
\item \mpnet~\cite{Qureshi2020motion}: This is the sampling-biasing version of Motion Planning Networks.
Given a training dataset of \threed point cloud workspaces, \xs, \xg, and solution paths, \mpnet learns to iteratively produce samples that mimic the solution paths.
We adapted the provided implementation and tuned its hyperparameters to achieve the best performance for the given problems. 
\item \flame~\cite{Chamzas2021}: This framework is similar to \fire and also retrieves \samples from a \DB.
    However, the local primitives are simpler, including only workspace information (\lw) and not considering \xg or \xs. The similarity function considers $\lw_i$ similar to $\lw_j$ if they have the same position and binary representation.
\item \static~\cite{Iversen2016Kernel, Lehner2017}: These methods generate a static sampling distribution by extracting key configurations from past trajectories.
    They do not rely on a similarity function but instead attempt to capture the problem's invariants.
    We emulate the static sampling idea of these methods by retrieving all the \cspace samples we have stored in \DB.
\end{itemize}

We consider these methods representative of the works discussed in~\autoref{sec:related}, with \mpnet being a non-retrieval method that directly maps \M to \cspace samples using a neural network, 
\flame a retrieval-based method with a hand-crafted similarity function, and \static a method that learns problem invariants.

We evaluate the performance of \fire and the generalization of the learned similarity function when both the training and testing examples come from the same dataset~(\autoref{sec:small}),
and also when the testing dataset is increasingly different from the training dataset~(\autoref{sec:test}). Finally, we evaluate \fire when retrieving experiences it was not trained on, and while the \DB includes unrelated experiences~(\autoref{sec:cage}). For our experiments we used Robowflex with MoveIt~\cite{Moveit, Robowflex} and the \textsc{ompl} benchmarking tools~\cite{Moll2015}.
The sampling parameters for \fire were the same as~\cite{Chamzas2021} \mbox{($\sigma^2 =0.2, \lambda =0.5$)}.

\subsection{Generalizing in similar problems}\label{sec:small}
\subsubsection{Learning (Training)}
In this experiment, \mpnet, \flame, and \fire were trained in problems that come from the \smalld dataset.
\fire and \flame were given enough training examples for their performance to converge in the \smalld dataset.
By convergence, we mean that the average planning time did not improve after doubling the number of experiences in \DB.
Specifically, \fire was trained with a total of 500 training examples. From these 500 examples, 200 were used to learn the similarity function and all of the 500 examples were added to \DB.
Training the Siamese network of \fire took around 1 hour for 200 epochs.  
\flame was trained with 1000 examples which were added to \DB as described in~\cite{Chamzas2021}.
Since it was difficult to profile the convergence of \mpnet ($\approx$1~day of training time) we provided it 5000 training examples to ensure that it has enough data.
This is of a similar order to~\cite{Qureshi2020motion} (10000).

\subsubsection{Evaluation (Testing)}
The methods were tested in a different set of 100 problems that also come from \smalld.
As seen in \autoref{fig:small}b, \fire outperformed all other methods in all four different settings in terms of planning time.
We do include the retrieval time in the total planning time for \flame and \fire but it was negligible in all cases \mbox{($0.01-0.1$ seconds)}.  
We also notice that the tuning of the underlying planner and the use of experiences interact synergistically, with the best performance being achieved by \fire with \rrtct.

\subsection{Generalizing in increasingly different problems}\label{sec:test}
\subsubsection{Learning (Training)}
We do not perform any additional training in these experiments and simply use the methods trained on \smalld from \autoref{sec:small}.

\subsubsection{Evaluation (Testing)}
In these experiments, the methods were tested on three datasets that are increasingly different from \smalld as shown in \autoref{fig:envs}.
The \talld is created by stacking the \smalld three times.
The \thind is also a bookcase but is different from \talld and \smalld because there is a divider and the distance between the shelves has changed.
Finally, \tabled is significantly different from \smalld regarding \W.
We used 100 testing examples for each of these three datasets.  
As shown in \autoref{fig:test}, \mpnet could not outperform \uni in \talld and \tabled except for \rrtcd, while in \thind it was not able to improve upon \uni given the time limits.
In some cases \mpnet performed worse than \uni; we attribute this behavior to the testing examples being outside the training dataset of \mpnet.
\flame did offer some improvement for the \talld environment but could not transfer to \thind or \tabled.
Also, in some cases \flame performed worse than \uni; this is attributed to the retrieval of very few critical samples leading to poor biased sampling (if nothing is retrieved it defaults to \uni).
On the other hand, \fire outperformed all other methods even in \thind and \tabled, demonstrating that the learned similarity function generalizes to problems that are significantly different than those in the training dataset.
We also note that \tabled has a different \xs configuration than the training dataset \smalld. This demonstrates the usefulness of independently considering \xs and \xg in the local primitives defined by \fire.

\begin{figure}
  \centering
  \includesvg[width=\linewidth]{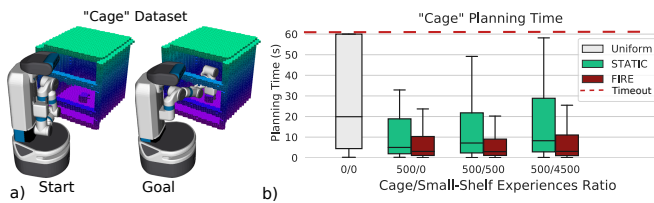}  
  \vspace{-1em}
  \caption{ 
      \textbf{a)} An example problem from the \caged dataset. 
      \textbf{b)} Planning time for 100 test examples from the \caged dataset using the \rrtct planner. The timeout was set to 60 seconds.
      The x-axis shows the number of experiences that exist in \DB from \smalld and from \caged. Note that \smalld and \caged have very different solution paths.
      In other words, the experiences from \smalld do not transfer to \caged. 
}\label{fig:cage}
\end{figure}

\subsection{Robustness to irrelevant experiences}\label{sec:cage}
\subsubsection{Learning (Training)}
In this experiment, we do not retrain \fire's similarity function and use the one obtained from training on \smalld from \autoref{sec:small}.
However, now we add to \DB example problems from both \caged and \smalld.
Note that the problems from \caged and \smalld are highly dissimilar in terms of solution paths. Thus, when solving a problem from \caged a good similarity function should not retrieve experiences generated from \smalld.
The x-axis in \autoref{fig:cage}b shows the ratio of example problems from \caged and \smalld.
For example, $500/0$ denotes an experience database \DB that has 500 examples from \caged and 0 examples from \smalld.

\subsubsection{Evaluation (Testing)}
In this experiment, we tested on 100 example problems from the \caged dataset using \rrtct as the underlying planner.  
We compared with \static to illustrate how irrelevant experiences from \smalld affect performance.
The results in \autoref{fig:cage}b show that although \static significantly outperforms \uni, its performance degrades as we add irrelevant experiences in the training dataset.   
On the other hand, \fire is robust to the irrelevant experiences from \smalld added to \DB since it maintains its good performance even with the $500/4500$ ratio.
\fire's similarity function was only trained on \smalld while \DB includes experiences from \caged.
This demonstrates that the learned latent space can successfully structure local primitives it was not trained on. 

\section{Conclusion}\label{sec:discussion}
In this work, we have proposed \fire, a framework that learns a similarity function for motion planning problems with sensed environments.
Using the learned similarity function, \fire retrieves relevant experiences from a database in the form of \samples that can informatively guide any sampling-based motion planner.
Through our experiments, we demonstrated the generalization of \fire outside its training dataset.
Furthermore, \fire can also learn incrementally without retraining by simply adding experiences in \DB, and can discriminate between relevant and irrelevant experiences.

In the future, we would like to improve \fire by bounding its memory requirements and treating biased samples differently from uniform samples \cite{Molina2020Link, Ichter2020LocalCrit}.
Additionally, we would like to investigate how the same ideas apply to other problems that include motion planning such as task and motion planning or kinodynamic planning.

\bibliographystyle{ieeetr}
\bibliography{references}

\end{document}